\pdfoutput=1
\documentclass[10pt]{article}

\usepackage[margin=1in]{geometry}
\usepackage{microtype}
\usepackage{amsmath,amssymb,amsfonts}
\usepackage{graphicx}
\usepackage{subcaption}
\usepackage{booktabs}
\usepackage{tabularx}
\usepackage{xcolor}
\usepackage{url}
\usepackage[numbers,sort&compress]{natbib}
\usepackage{hyperref}
\hypersetup{
  colorlinks=true,
  linkcolor=blue,
  citecolor=blue,
  urlcolor=blue
}

\usepackage{tikz}
\usetikzlibrary{positioning,arrows.meta}

\usepackage{float}

\title{SpatialBench-UC: Uncertainty-Aware Evaluation of Spatial Prompt Following in Text-to-Image Generation}
\author{Amine Rostane\\{\small ESIEA, France}\\{\small\texttt{hello@aminerostane.com}}}
\date{}

\begin{document}
\maketitle

\begin{abstract}
\noindent\setlength{\parindent}{0pt}%
Evaluating whether text to image models follow explicit spatial instructions is difficult to automate. Object detectors may miss targets or return multiple plausible detections, and simple geometric tests can become ambiguous in borderline cases. Spatial evaluation is naturally a \textbf{selective prediction} problem, a checker should be allowed to abstain when evidence is weak and should report confidence so results can be interpreted as a risk--coverage trade-off rather than a single score.

We introduce \textbf{SpatialBench-UC}, a small, reproducible benchmark for pairwise spatial relations. The benchmark contains 200 prompts (50 object pairs $\times$ 4 relations) grouped into 100 counterfactual pairs obtained by swapping object roles (e.g., \emph{A left of B} $\leftrightarrow$ \emph{B right of A}). We release a benchmark package, versioned prompts, pinned configs, per-sample checker outputs, and report tables, enabling reproducible and auditable comparisons across models. We also include a lightweight human audit (N=200) used to calibrate the checker’s abstention margin and confidence threshold.

On three baselines, Stable Diffusion 1.5 (prompt-only), SD~1.5+BoxDiff, and SD~1.4+GLIGEN, the checker reports (PASS rate / coverage): 11.8\% / 23.8\%, 40.4\% / 42.5\%, and 51.6\% / 52.0\%, with conditional PASS rates of 49.5\%, 95.0\%, and 99.3\% on decided samples. \textbf{PASS denotes the checker’s judgment}, not human-verified accuracy unless validated by audit labels.

\end{abstract}

\section{Introduction}
Text-to-image diffusion models can produce visually compelling images, yet they often violate explicit spatial constraints stated in text (e.g., \emph{``a dog to the left of a chair''}). Scoring spatial prompt following reliably is harder than it looks, automated evaluation typically depends on intermediate perception (object detection), which introduces common uncertainty sources, missed objects, multiple plausible instances, and ambiguous geometry when objects overlap or lie near relation boundaries. Collapsing these uncertainties into a single scalar score can make comparisons hard to interpret and hard to reproduce.

Spatial evaluation is naturally a \textbf{selective prediction} problem, an evaluator should output PASS/FAIL when evidence is sufficient, abstain otherwise, and attach confidence so users can trade coverage for lower risk. Throughout, PASS/FAIL/UNDECIDABLE are \emph{checker verdicts} (PASS is not ground-truth accuracy unless validated by human labels).

\paragraph{Contributions.}
\begin{itemize}
  \item \textbf{Uncertainty-aware evaluation framework.}
  We formalize spatial evaluation with explicit abstention (PASS/FAIL/UNDECIDABLE) and interpretable confidence scores.

  \item \textbf{Reproducible benchmark package.}
  We release versioned prompts with hashes, pinned configurations, per-sample evaluation outputs, and structured metadata to support independent replication and auditing.

  \item \textbf{Counterfactual prompts pairing.}
  The benchmark comprises 200 prompts (50 object pairs $\times$ 4 axis-aligned relations) organized into 100 counterfactual pairs via role swapping (e.g., \emph{A left of B} $\leftrightarrow$ \emph{B right of A}).

  \item \textbf{Human audit and calibration protocol.}
  A lightweight audit (N=200) grounds risk--coverage analysis and calibrates the abstention margin and confidence threshold.

  \item \textbf{Empirical comparison of generation strategies.}
  We evaluate SD~1.5 (prompt-only), SD~1.5 with BoxDiff, and GLIGEN (SD~1.4) on a fixed set of generated images.
\end{itemize}
See Sections~\ref{sec:audit_calibration}--\ref{sec:results} for the audit protocol and results; Figure~\ref{fig:risk_coverage} and Table~\ref{tab:abstention_breakdown} summarize the selective prediction and abstention behavior.

\begin{figure}[H]
  \centering
  \begin{subfigure}[t]{0.32\linewidth}
    \centering
    \includegraphics[width=\linewidth]{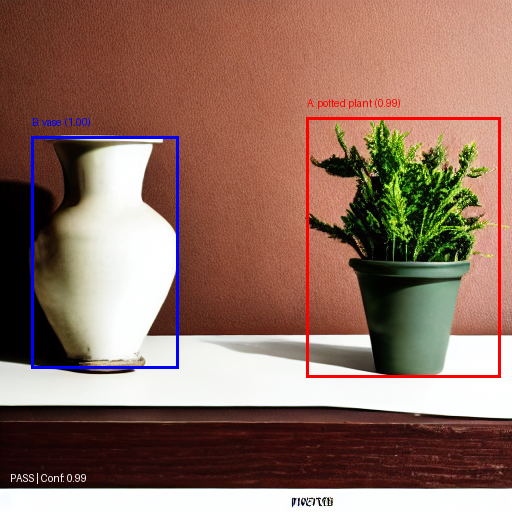}
    \caption{\textbf{PASS.}\\\footnotesize Prompt: \emph{potted plant right of vase}}
  \end{subfigure}\hfill
  \begin{subfigure}[t]{0.32\linewidth}
    \centering
    \includegraphics[width=\linewidth]{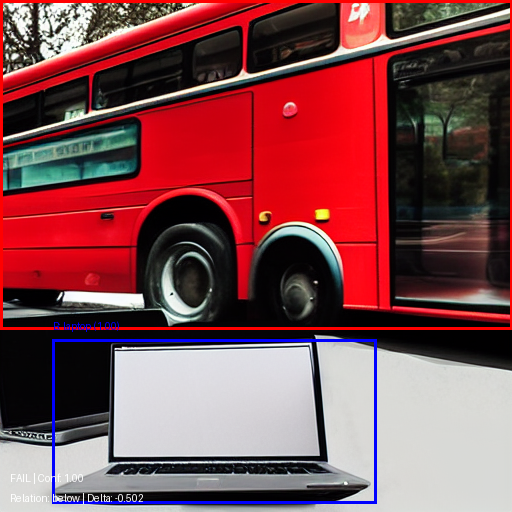}
    \caption{\textbf{FAIL.}\\\footnotesize Prompt: \emph{bus below laptop}}
  \end{subfigure}\hfill
  \begin{subfigure}[t]{0.32\linewidth}
    \centering
    \includegraphics[width=\linewidth]{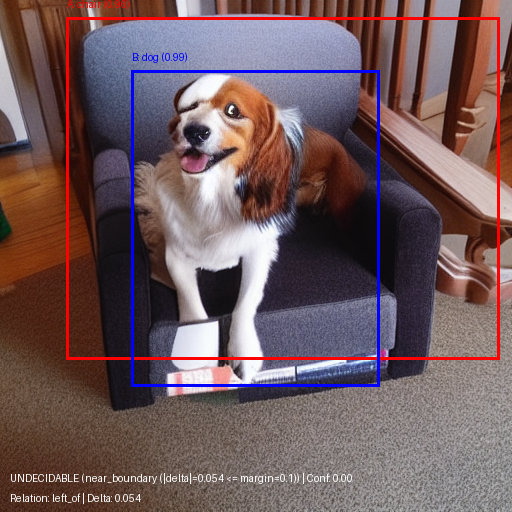}
    \caption{\textbf{UNDECIDABLE (near-boundary).}\\\footnotesize Prompt: \emph{chair left of dog}}
  \end{subfigure}
  \caption{\textbf{Qualitative examples of checker outcomes.} Each panel shows a generated image with detector boxes (red=A, blue=B) and the checker verdict. We include one clear PASS, one clear FAIL, and one abstention example where geometry is near the decision boundary.}
  \label{fig:teaser}
\end{figure}

\section{Evaluation under uncertainty}
Uncertainty is the central obstacle in spatial evaluation. Rather than forcing every sample into PASS or FAIL, we allow the evaluator to abstain (UNDECIDABLE) when the available evidence is insufficient. This section characterizes the sources of ambiguity that motivate abstention and clarifies the scope of what SpatialBench-UC measures.

\subsection{Sources of evaluation uncertainty}
Spatial prompt following is easy to state but difficult to verify automatically. Even when restricting attention to axis-aligned relations between two objects, several sources of uncertainty arise. First, \textbf{missing detections}, object detectors may fail to locate one or both targets, either because the objects are absent, too small, or simply missed by the model. Second, \textbf{ambiguous referents}, when multiple instances of the same object class appear in an image, identifying which pair corresponds to the prompt becomes unclear. Third, \textbf{overlaps and near-boundary cases}, geometric relations can become ill-defined when objects overlap substantially or when their bounding-box centers lie near the decision boundary separating ``left'' from ``right'' or ``above'' from ``below.'' Finally, \textbf{perturbation sensitivity}, small image changes such as blur, brightness shifts, or resizing can alter detection outputs or flip the geometric verdict, revealing instability in the evaluation pipeline itself.

These ambiguities are not rare edge cases. In our fixed evaluation set, UNDECIDABLE outcomes are common, and the dominant causes are directly measurable (Table~\ref{tab:abstention_breakdown}).

\subsection{Evaluator semantics and scope}
SpatialBench-UC reports \textbf{checker verdicts} (PASS, FAIL, or UNDECIDABLE) together with a \textbf{confidence score} for each sample. A PASS verdict indicates that the detector--geometry checker judged the spatial relation to be satisfied; \textbf{it should not be interpreted as ground-truth accuracy} unless corroborated by human labels (Section~\ref{sec:audit_calibration}).

The benchmark measures compliance with axis-aligned spatial relations \emph{as judged by the evaluator}, given pretrained object detectors and explicit abstention rules. Consequently, reported results reflect both the generator's spatial prompt-following ability and the detectability of the target objects under the chosen detectors. The benchmark does not establish ground-truth correctness for every image, does not cover relations beyond the four axis-aligned cases, and does not claim that the confidence score constitutes a calibrated probability.

\subsection{Uncertainty sources handled by the evaluator}
Table~\ref{tab:uncertainty_sources} summarizes the main uncertainty sources and how they map to checker behavior. This table appears before pipeline details to make abstention feel like an explicit, inspectable design choice rather than a black box.

\begin{table}[H]
  \centering
  \begin{tabularx}{\linewidth}{@{}l>{\raggedright\arraybackslash}X@{}}
\toprule
Uncertainty source & How the evaluator handles it (decision + implication) \\
\midrule
Missing detections &
If either object is not detected after thresholding/size filtering, the checker abstains (\mbox{UNDECIDABLE}; reason \texttt{missing}) and sets confidence to 0. \\
\addlinespace
Multiple instances / ambiguous referents &
If multiple instances of the target label have similar scores (within \texttt{ambiguity\_delta}), the checker abstains (\mbox{UNDECIDABLE}; reason \texttt{ambiguous}) and sets confidence to 0. \\
\addlinespace
Strong overlap between boxes (horizontal relations) &
For \texttt{left\_of}/\texttt{right\_of}, if IoU exceeds \texttt{max\_overlap\_iou}, the checker abstains (\mbox{UNDECIDABLE}; reason \texttt{high\_overlap}) because center ordering is unreliable under heavy overlap. \\
\addlinespace
Near-threshold geometry &
If the normalized center delta lies within a margin band ($|d|\le m$), the checker abstains (\mbox{UNDECIDABLE}; reason \texttt{near\_boundary}); geometry confidence goes to (near) zero by construction. \\
\addlinespace
Sensitivity to perturbations &
If a decided verdict (PASS/FAIL) is unstable under lightweight perturbations, the checker abstains (\mbox{UNDECIDABLE}; reason \texttt{unstable}) and the stability component penalizes confidence. Under the current perturbations/threshold, this mechanism did not activate in our runs. \\
\bottomrule
\end{tabularx}

  \caption{Uncertainty sources explicitly handled by the evaluator and their implications for PASS/FAIL/UNDECIDABLE and confidence.}
  \label{tab:uncertainty_sources}
\end{table}

\paragraph{Note on stability in the current runs.}
Stability is implemented and contributes to confidence; with our current perturbation set and threshold, it did not cause additional abstentions in these runs (Table~\ref{tab:abstention_breakdown}). We keep it as a configurable safeguard for settings where detector/geometry sensitivity is higher.

\section{Related work}
We place SpatialBench-UC in the context of prior spatial benchmarks for text-to-image generation and of evaluation under abstention. Our emphasis is on detector+geometry verification because it is transparent and auditable, but this choice makes detector dependence central to interpretability (Section~\ref{sec:discussion}).

\subsection{Spatial evaluation for text-to-image generation}
Spatial relationship evaluation for text-to-image generation is commonly operationalized by running object detectors and applying geometric checks on detected boxes. SR2D/VISOR introduced a spatial prompt benchmark and automatic verification built on detected object locations \citep{gokhale2022benchmarking}. GenEval includes a \emph{position} category evaluated through detection and geometry \citep{ghosh2023geneval}. T2I-CompBench provides a broader compositional benchmark that includes spatial relations as a subset \citep{huang2023t2icompbench}. SpatialBench-UC follows this lineage, but makes uncertainty explicit by exposing abstention reasons and confidence and by reporting coverage alongside checker PASS rates.

Detector+geometry evaluation is not the only option. TIFA evaluates text-to-image faithfulness by converting prompts into question--answer pairs and using VQA models, providing a complementary, interpretable approach \citep{Hu_2023_ICCV}. Counterfactual pairing is also related to minimal-pair evaluation in vision--language benchmarks such as Winoground \citep{Thrush_2022_CVPR}; we adapt the idea to spatial relations by pairing logically equivalent prompts (e.g., \emph{A left of B} $\leftrightarrow$ \emph{B right of A}) and reporting pair-level outcomes and undecidable mass.

For a demonstration study we compare prompt-only Stable Diffusion (latent diffusion) \citep{Rombach_2022_CVPR} to generation methods that incorporate explicit spatial grounding. BoxDiff applies training-free box-constrained diffusion via attention manipulation \citep{xie2023boxdiff}, and GLIGEN enables grounded generation conditioned on bounding boxes \citep{li2023gligen}. We cite ControlNet as an example of conditional control for diffusion models \citep{Zhang_2023_ICCV_ControlNet}.

\subsection{Selective prediction and calibration}
Our abstention and risk--coverage reporting connect to classical and modern work on selective prediction. The reject option formalizes an optimal error--reject trade-off \citep{chow1970optimum}. Selective classification for deep networks studies risk as a function of coverage \citep{GeifmanElYaniv2017Selective}. Calibration work motivates audit-driven selection of confidence operating points \citep{guo2017calibration}, though we do not claim probabilistic calibration of our confidence score.

\section{Benchmark and artifact package}
This section defines the benchmark instance and the released materials used throughout the paper. We evaluate on a fixed set of generated images so evaluation and reporting can be rerun without resynthesizing images.

\subsection{Prompts and counterfactual pairs (v1.0.1)}
SpatialBench-UC Prompts v1.0.1 contains 200 English prompts describing two objects and one spatial relation from \{\texttt{left\_of}, \texttt{right\_of}, \texttt{above}, \texttt{below}\}. The dataset is built from 50 \emph{unordered object pairs} expanded into four directional relations (50 pairs $\times$ 4 relations = 200 prompts), using a simple photographic template (e.g., ``A photo of a cat to the left of a chair.'') to reduce stylistic variation. These 200 prompts are additionally grouped into 100 \emph{counterfactual pairs}: for each unordered object pair $(A,B)$, we form one left/right pair and one above/below pair by swapping roles.

Each prompt is paired with a counterfactual prompt that is logically equivalent under role swapping:
\begin{itemize}
  \item \emph{A left of B} $\leftrightarrow$ \emph{B right of A}
  \item \emph{A above B} $\leftrightarrow$ \emph{B below A}
\end{itemize}
This pairing enables pair-level analysis of consistency and abstention mass (Section~\ref{sec:results}).

\begin{table}[H]
  \centering
  \begin{tabular}{@{}ll@{}}
    \toprule
    Total prompts & 200 \\
    Object pairs & 50 \\
    Counterfactual pairs & 100 \\
    Relations & 4 (\texttt{left\_of}, \texttt{right\_of}, \texttt{above}, \texttt{below}) \\
    Prompts per relation & 50 \\
    Unique objects & 33 \\
    \bottomrule
  \end{tabular}
  \caption{SpatialBench-UC Prompts v1.0.1 summary (from dataset metadata and prompt file).}
  \label{tab:dataset_stats}
\end{table}

\subsection{Fixed evaluation set (three methods, K=4 seeds)}
We evaluate three generation strategies that differ in the amount of explicit spatial grounding:
\begin{itemize}
  \item \textbf{SD~1.5 prompt-only}: standard text-to-image generation \citep{Rombach_2022_CVPR}.
  \item \textbf{SD~1.5 BoxDiff}: training-free box-constrained diffusion via attention manipulation \citep{xie2023boxdiff}.
  \item \textbf{GLIGEN (SD~1.4)}: grounded generation conditioned on bounding boxes \citep{li2023gligen}.
\end{itemize}
Each prompt is generated with $K=4$ seeds, producing 800 images per method (200 prompts $\times$ 4 seeds), at 512$\times$512 resolution with 30 diffusion steps and guidance scale 7.5. Model IDs and revisions are pinned in the generator configs (Appendix~\ref{app:repro_paths}).

\subsection{Release package and pipeline}
\begin{figure}[H]
  \centering
  \resizebox{\linewidth}{!}{%
    \begin{tikzpicture}[
      box/.style={draw, rounded corners, align=center, inner sep=4pt},
      arrow/.style={-Latex, thick}
    ]
      \node[box] (prompts) {Prompts\\(versioned + hashed)};
      \node[box, right=6mm of prompts] (gen) {Generate\\(fixed images)};
      \node[box, right=6mm of gen] (detect) {Detect\\(primary + secondary)};
      \node[box, right=6mm of detect] (check) {Relation check\\+ abstain};
      \node[box, right=6mm of check] (conf) {Confidence\\(det/geom/stab/agree)};
      \node[box, right=6mm of conf] (report) {Report\\(tables + plots)};
      \node[box, below=8mm of report] (audit) {Audit \& calibrate\\(human labels)};
      \node[box, left=6mm of audit] (cfg) {Checker config\\(calibrated)};

      \draw[arrow] (prompts) -- (gen);
      \draw[arrow] (gen) -- (detect);
      \draw[arrow] (detect) -- (check);
      \draw[arrow] (check) -- (conf);
      \draw[arrow] (conf) -- (report);
      \draw[arrow] (report) -- (audit);
      \draw[arrow] (audit) -- (cfg);
    \end{tikzpicture}%
  }
  \caption{Pipeline overview. The evaluator outputs PASS/FAIL/UNDECIDABLE with confidence, enabling reporting under abstention; a small human audit calibrates parameters and operating points.}
  \label{fig:pipeline}
\end{figure}
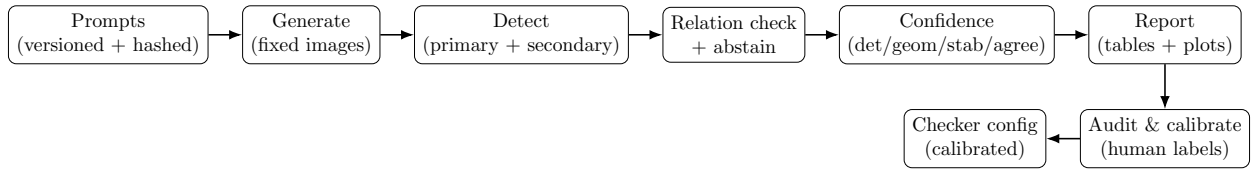

We release prompts, configs, per-sample checker outputs, and report tables so analyses can be reproduced and inspected from the fixed runs. Figure~\ref{fig:pipeline} summarizes the end-to-end pipeline from prompts to calibrated reporting. Exact paths are listed in Appendix~\ref{app:repro_paths}.

\section{Uncertainty-aware evaluator and metrics}
We now specify the evaluator and metrics with an emphasis on transparency, each verdict is grounded in detected boxes and explicit geometric rules, with abstention and decomposed confidence. This keeps decisions auditable, but necessarily inherits detector limitations.

\subsection{Checker: detectors, decision rule, and abstention}
Given an image and a prompt specifying objects $(A,B)$ and relation $r \in \{\texttt{left\_of},\texttt{right\_of},\texttt{above},\texttt{below}\}$, the checker outputs a verdict in \{\textbf{PASS}, \textbf{FAIL}, \textbf{UNDECIDABLE}\} and a confidence score in $[0,1]$.

\paragraph{Detectors.}
We use a closed-vocabulary COCO detector (Faster R-CNN) \citep{ren2015faster,LinMBHPRDZ14} and an open-vocabulary detector (Grounding DINO) \citep{liu2023grounding}. The secondary detector is used only to form an agreement signal; it does not override the primary verdict.

\paragraph{Detection selection.}
For each target label, we filter detections by a score threshold and minimum area fraction and select the highest-scoring remaining box. If either object is missing (\texttt{missing}) or if multiple instances are ambiguous (\texttt{ambiguous}), the checker abstains (UNDECIDABLE) and assigns confidence 0 (Table~\ref{tab:uncertainty_sources}).

\paragraph{Geometry and near-boundary abstention.}
Let $(c_x(A),c_y(A))$ and $(c_x(B),c_y(B))$ denote bounding box centers, and let $W,H$ be image width/height. We define:
\[
  d_x = \frac{c_x(A)-c_x(B)}{W}, \qquad
  d_y = \frac{c_y(A)-c_y(B)}{H}.
\]
For horizontal relations we use $d_x$ and for vertical relations we use $d_y$. With margin $m$, the checker abstains if $|d|\le m$ (\texttt{near\_boundary}). Outside the margin, PASS/FAIL are determined by the expected sign of $d$ (e.g., for \texttt{left\_of}, PASS if $d_x<-m$; for \texttt{right\_of}, PASS if $d_x>m$).

\paragraph{Overlap and stability.}
For \texttt{left\_of}/\texttt{right\_of}, if the selected boxes overlap too strongly (IoU above a threshold), the checker abstains (\texttt{high\_overlap}). If the initial verdict is decided, we test stability under lightweight perturbations (brightness/blur/resize); if stability drops below a threshold, we abstain (\texttt{unstable}).

\paragraph{Key parameter values.}
For concreteness and reproducibility, Table~\ref{tab:checker_config} lists the key calibrated checker parameters used for the calibrated report.

\begin{table}[H]
  \centering
  \begin{tabularx}{\linewidth}{@{}ll>{\raggedright\arraybackslash}X@{}}
\toprule
Component & Parameter & Value \\
\midrule
Thresholds & \texttt{detection\_score} & 0.2 \\
Thresholds & effective detection cutoff (primary) & $\max(t_{\mathrm{det}}, t_{\mathrm{detector}})=\max(0.2, 0.5)=0.5$ \\
Thresholds & \texttt{min\_area\_fraction} & 0.005 \\
Thresholds & \texttt{ambiguity\_delta} & 0.1 \\
Thresholds & \texttt{max\_overlap\_iou} & 0.5 \\
Geometry & \texttt{margin} ($m$) & 0.1 \\
Stability & \texttt{consistency\_threshold} & 0.5 \\
Stability & perturbations & brightness (1.1, 0.9), blur ($\sigma{=}1.0$), resize (0.9) \\
Confidence & \texttt{geom\_slope} ($\gamma$) & 0.15 \\
Confidence & weights $(w_d,w_g,w_s,w_a)$ & (0.4, 0.3, 0.2, 0.1) \\
Primary detector & Faster R-CNN \texttt{score\_threshold} & 0.5 \\
Secondary detector & GroundingDINO \texttt{box\_threshold} / \texttt{text\_threshold} & 0.35 / 0.25 \\
\bottomrule
\end{tabularx}

  \caption{Key calibrated checker parameters (from \texttt{configs/checker\_v1.yaml}).}
  \label{tab:checker_config}
\end{table}

\subsection{Confidence score}
Confidence is designed to be interpretable rather than probabilistic. We compute four components, detection strength, distance to the margin, stability, and detector agreement. The overall confidence is a weighted geometric mean:
\[
  \mathrm{Conf} = (\mathrm{det}+\varepsilon)^{w_d} (\mathrm{geom}+\varepsilon)^{w_g} (\mathrm{stab}+\varepsilon)^{w_s} (\mathrm{agree}+\varepsilon)^{w_a}.
\]
Detection confidence is $\sqrt{s_A s_B}$ from primary detector scores; geometry confidence is $\mathrm{clip}((|d|-m)/\gamma,0,1)$; stability and agreement are in $[0,1]$. When the secondary detector fails, we set agreement to 0.5 (neutral) rather than $0$ (punitive); a missing secondary signal is uninformative, not evidence against the primary decision.

\subsection{Metrics under abstention}
We report both per-image and prompt-level metrics computed on checker outputs. Since the checker can abstain, we always report \emph{coverage} alongside PASS rates. Throughout, \textbf{PASS/FAIL/UNDECIDABLE are checker verdicts} (PASS is not ground-truth accuracy unless validated by human labels).

\paragraph{Per-image metrics.}
For $N$ images with counts $(N_P, N_F, N_U)$ for PASS/FAIL/UNDECIDABLE:
\[
  \text{pass\_rate} = \frac{N_P}{N}, \quad
  \text{coverage} = \frac{N_P+N_F}{N}, \quad
  \text{pass\_rate\_cond} = \frac{N_P}{N_P+N_F}.
\]

\paragraph{Prompt-level metrics.}
Each prompt has $K=4$ images (seeds). Best-of-$K$ prompt PASS is optimistic (PASS if any seed PASSes), so we report it as an upper bound. We also report all-of-$K$ prompt PASS (PASS only if all seeds PASS), which is stricter and captures seed sensitivity (Table~\ref{tab:prompt_metrics}).

\paragraph{Counterfactual consistency.}
Prompts are paired into counterfactual equivalents. For each pair we report both-pass rate and undecidable mass; one-sided contradictions are reported explicitly when present.

\paragraph{Risk--coverage on audited subset.}
On the human-audited subset, checker confidence serves as a selective prediction score; we sweep a threshold $\tau$. Coverage is computed over all audited samples, while accuracy/risk exclude human-UNDECIDABLE labels from the accuracy denominator (Section~\ref{sec:audit_calibration}).

\section{Human audit and calibration}\label{sec:audit_calibration}
A small human audit anchors the evaluator, it defines how risk is computed on decided labels and selects operating parameters via an explicit objective. This strengthens interpretability of confidence and abstention, while remaining limited by audit size and single-annotator labeling.

\subsection{Audit protocol and sampling}
We sample $N=200$ images for human auditing using stratified sampling across \textbf{method} $\times$ \textbf{relation} $\times$ \textbf{checker verdict}, and (when possible) across confidence bins. Human labels are PASS/FAIL/UNDECIDABLE. UNDECIDABLE is used when the relation cannot be determined confidently (e.g., missing/unclear objects, heavy overlap, or near-ties in spatial ordering).

\paragraph{Risk definition and human-UNDECIDABLE.}
Risk--coverage is computed on the audited subset using the checker confidence as a selection score. Coverage counts all audited samples, but accuracy/risk are computed only on covered samples with human PASS/FAIL labels; samples labeled UNDECIDABLE by humans are excluded from the accuracy denominator rather than being forced into PASS/FAIL.

\subsection{Calibration objective and selected parameters}
We perform a grid search over margin $m$, detection threshold $t_{\mathrm{det}}$, and confidence threshold $\tau$ (Table~\ref{tab:calibration_params}). For each parameter triple we re-evaluate audited samples and minimize:
\[
  J(m,t_{\mathrm{det}},\tau)=10\cdot \mathrm{FPR}^{\mathrm{PASS}}_{\ge\tau}+2\cdot \mathrm{Risk}(\tau)+0.5\cdot(1-\mathrm{Coverage}(\tau)),
\]
where $\mathrm{FPR}^{\mathrm{PASS}}_{\ge\tau}$ is the fraction of checker PASS predictions with confidence $\ge\tau$ that humans label FAIL, $\mathrm{Coverage}(\tau)$ is the fraction of audited samples covered by the checker (PASS/FAIL with confidence $\ge\tau$), and $\mathrm{Risk}(\tau)=1-\mathrm{Accuracy}(\tau)$ with accuracy computed on human PASS/FAIL labels only.
Because detector wrappers apply their own score thresholds (e.g., our Faster R-CNN wrapper discards detections below 0.5), the effective detection cutoff is $\max(t_{\mathrm{det}}, t_{\mathrm{detector}})$; thus setting $t_{\mathrm{det}}<0.5$ does not change Faster R-CNN outputs in the released configuration.

\begin{table}[H]
  \centering
  \begin{tabular}{@{}lll@{}}
\toprule
Parameter & Searched values & Selected \\
\midrule
Margin $m$ & \{0.03, 0.05, 0.07, 0.10\} & 0.10 \\
Detection threshold $t_{\mathrm{det}}$ & \{0.2, 0.3, 0.4\} & 0.20 \\
Confidence threshold $\tau$ & \{0.3, 0.5, 0.7\} & 0.70 \\
\bottomrule
\end{tabular}

  \caption{Calibration grid and selected parameters (from the audit-driven search).}
  \label{tab:calibration_params}
\end{table}

The selected parameters are $m=0.1$, $t_{\mathrm{det}}=0.2$, and $\tau=0.7$. Figure~\ref{fig:risk_coverage} reports the risk--coverage curve on the audited subset for the \emph{calibrated} checker (Appendix Figure~\ref{fig:risk_coverage_uncal} shows the uncalibrated curve).

\section{Results}\label{sec:results}
We report results under abstention, coverage is reported alongside checker PASS rates, and risk--coverage is computed on audited labels. Some quantities are conditional by design (e.g., PASS$\mid$Decided); we make these conditionals explicit and interpret them as evidence about decidability versus correctness.

\subsection{Main quantitative results (calibrated report)}
Table~\ref{tab:main_results} summarizes per-image checker metrics on the fixed evaluation set. PASS/FAIL/UNDECIDABLE are checker outputs, not ground-truth labels unless validated by human audit (Section~\ref{sec:audit_calibration}).

\begin{table}[H]
  \centering
  \begin{tabular}{@{}lrrrr@{}}
\toprule
Method & PASS (\%) & Coverage (\%) & PASS$\mid$Decided (\%) & MeanConf \\
\midrule
SD1.5 Prompt-only & 11.8 & 23.8 & 49.5 & 0.206 \\
SD1.5 BoxDiff & 40.4 & 42.5 & 95.0 & 0.395 \\
SD1.4 GLIGEN & 51.6 & 52.0 & 99.3 & 0.506 \\
\bottomrule
\end{tabular}

  \caption{Main results (calibrated report). PASS/FAIL/UNDECIDABLE are checker verdicts. Coverage is the fraction of images where the checker makes a decision (PASS/FAIL).}
  \label{tab:main_results}
\end{table}

\subsection{Selective prediction: risk--coverage on audited labels}
\begin{figure}[H]
  \centering
  \includegraphics[width=0.92\linewidth]{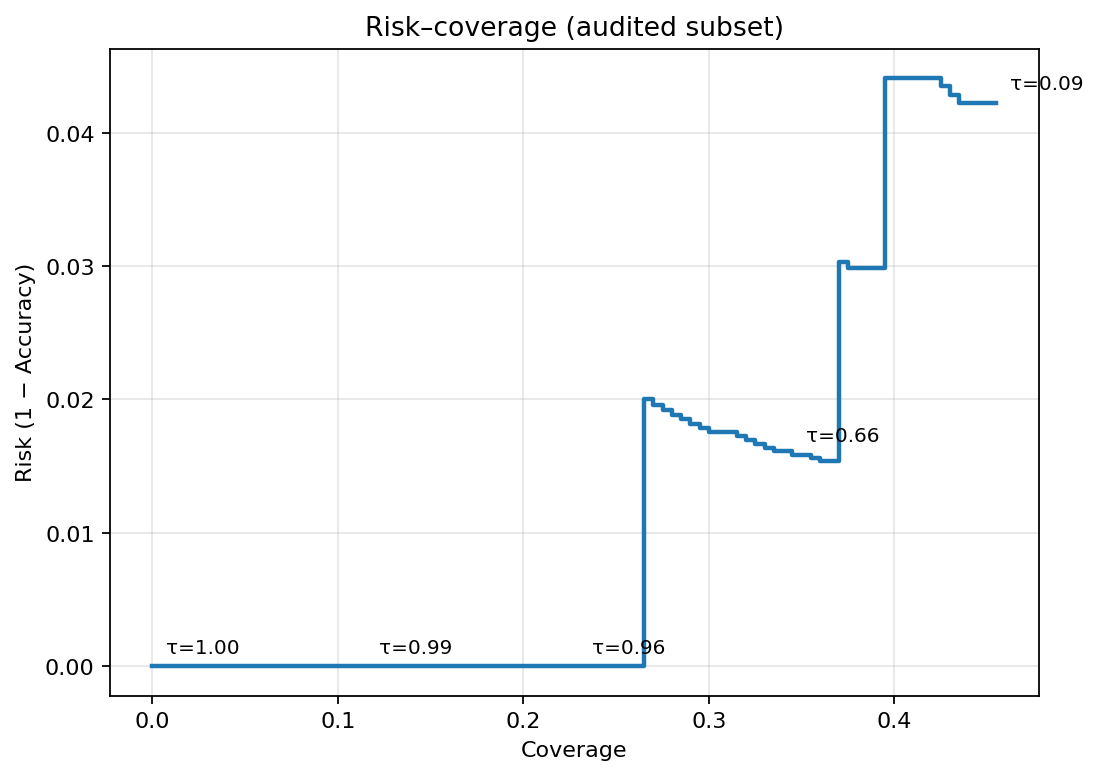}
  \caption{Risk--coverage on the audited subset for the calibrated checker (sweep over unique confidence values). The curve is step-like because coverage changes only when $\tau$ crosses a sample’s discrete confidence value. Higher confidence thresholds reduce risk but also reduce coverage; risk excludes human-UNDECIDABLE labels from the accuracy denominator.}
  \label{fig:risk_coverage}
\end{figure}

Figure~\ref{fig:risk_coverage} shows risk--coverage behavior on the audited subset when thresholding on checker confidence. Coverage counts all audited samples; accuracy/risk are computed only on covered samples with human PASS/FAIL labels (human-UNDECIDABLE are excluded from the accuracy denominator).

\begin{figure}[H]
  \centering
  \includegraphics[width=0.92\linewidth]{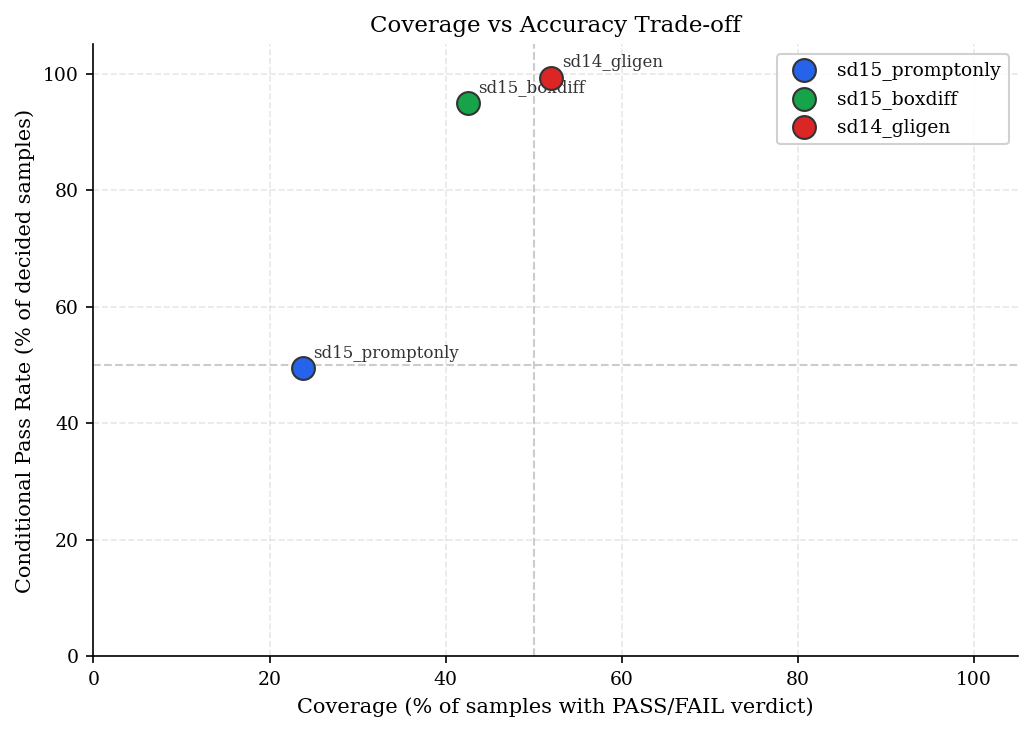}
  \caption{Coverage vs conditional PASS (calibrated report). Higher coverage means fewer abstentions; conditional PASS is computed only on decided samples (PASS/FAIL).}
  \label{fig:coverage_accuracy}
\end{figure}

\paragraph{Decidability versus conditional correctness.}
Figure~\ref{fig:coverage_accuracy} summarizes the tradeoff between \emph{decidability} (coverage) and conditional correctness on decided samples (PASS$\mid$Decided). This view complements Table~\ref{tab:main_results} by separating improvements due to fewer abstentions from improvements conditional on a decision.

\subsection{Abstention breakdown}
Table~\ref{tab:abstention_breakdown} breaks down UNDECIDABLE outcomes by reason. Missing detections dominate abstention across methods, which motivates interpreting results as a combination of spatial compliance and detectability under the chosen detectors (Section~\ref{sec:discussion}).

\begin{table}[H]
  \centering
  \resizebox{\linewidth}{!}{\begin{tabular}{@{}lrrrrrr@{}}
\toprule
Method & Undecidable (\%) & Missing (\%) & Ambiguous (\%) & Overlap (\%) & Near-boundary (\%) & Unstable (\%) \\
\midrule
SD1.5 Prompt-only & 76.25 & 56.00 & 9.12 & 0.25 & 10.88 & 0.00 \\
SD1.5 BoxDiff & 57.50 & 45.75 & 7.88 & 0.75 & 3.12 & 0.00 \\
SD1.4 GLIGEN & 48.00 & 38.25 & 9.62 & 0.00 & 0.12 & 0.00 \\
\bottomrule
\end{tabular}

}
  \caption{UNDECIDABLE breakdown by reason (percent of all images). Values sum to the UNDECIDABLE rate for each method.}
  \label{tab:abstention_breakdown}
\end{table}

\subsection{Prompt-level metrics: best-of-K vs all-of-K}
Table~\ref{tab:prompt_metrics} reports prompt-level PASS rates under both best-of-4 (optimistic upper bound) and all-of-4 (strict seed-robustness) definitions. For reproducibility, these prompt-level rates are also exported in the calibrated report tables (\texttt{tables/prompt\_metrics.csv}).

\begin{table}[H]
  \centering
  \begin{tabular}{@{}lrr@{}}
\toprule
Method & Best-of-4 PromptPass (\%) & All-of-4 PromptPass (\%) \\
\midrule
SD1.5 Prompt-only & 34.0 & 1.0 \\
SD1.5 BoxDiff & 76.0 & 8.5 \\
SD1.4 GLIGEN & 78.5 & 21.5 \\
\bottomrule
\end{tabular}

  \caption{Prompt-level PASS rates (calibrated report). Best-of-4 is optimistic; all-of-4 is strict.}
  \label{tab:prompt_metrics}
\end{table}

\subsection{Qualitative evidence}
\begin{figure}[H]
  \centering
  \begin{subfigure}[t]{0.32\linewidth}
    \centering
    \includegraphics[width=\linewidth]{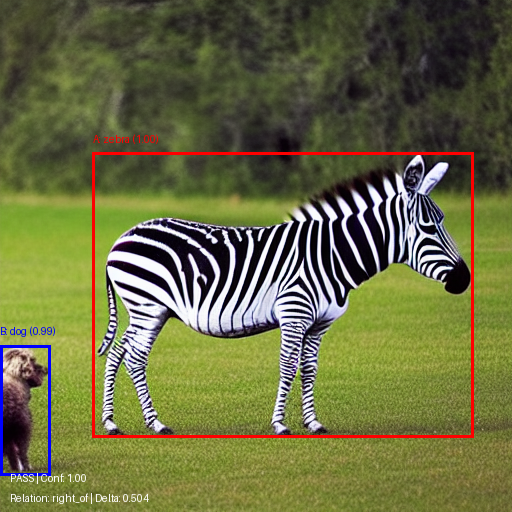}
    \caption{Prompt-only (audited PASS)\\``zebra right of dog''}
  \end{subfigure}\hfill
  \begin{subfigure}[t]{0.32\linewidth}
    \centering
    \includegraphics[width=\linewidth]{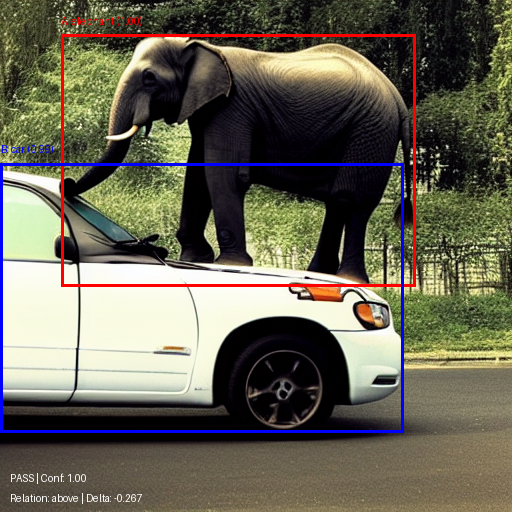}
    \caption{BoxDiff (audited PASS)\\``elephant above car''}
  \end{subfigure}\hfill
  \begin{subfigure}[t]{0.32\linewidth}
    \centering
    \includegraphics[width=\linewidth]{figures/overlay_gligen_success.png}
    \caption{GLIGEN (audited PASS)\\``potted plant right of vase''}
  \end{subfigure}
  \caption{Qualitative support: high-confidence PASS examples. Overlays show detected boxes and the checker verdict for the specified spatial relation.}
  \label{fig:qualitative_support}
\end{figure}

Figure~\ref{fig:qualitative_support} shows high-confidence checker PASS examples (audited PASS) for each method. These overlays illustrate what the evaluator considers a decided success, and complement the quantitative results.

\paragraph{Additional breakdowns and uncalibrated results.}
By-relation and counterfactual tables, as well as the uncalibrated report and calibration deltas, are provided in Appendix~\ref{app:additional_results}. A larger qualitative gallery of representative incorrect decisions and abstentions is also included in the appendix (Appendix~\ref{app:qual_gallery}).

\section{Discussion and conclusion}\label{sec:discussion}
Two takeaways stand out. First, grounding methods (BoxDiff, GLIGEN) substantially increase checker PASS among decided cases relative to prompt-only generation. Second, abstention remains a dominant factor in automated spatial evaluation, with missing detections accounting for most UNDECIDABLE outcomes (Table~\ref{tab:abstention_breakdown}).

\subsection{Discussion}
Across all reported metrics, explicit spatial grounding (BoxDiff, GLIGEN) substantially improves checker PASS rates relative to prompt-only generation, and also improves coverage (more cases where the evaluator can decide). This aligns with the intuition that grounding reduces both relational ambiguity and missing-object outcomes.

Calibration shifts the evaluator toward a safer operating regime, increasing the near-boundary margin and selecting a confidence threshold reduces risk at the cost of lower coverage, consistent with selective prediction behavior (Figure~\ref{fig:risk_coverage}).

Counterfactual evaluation emphasizes pair coverage (decidability) in addition to both-pass rate, on this benchmark, contradictions are rare relative to undecidable mass (Appendix~\ref{app:additional_results}).

\subsection{Limitations}
The evaluator depends on pretrained detectors and therefore inherits detector limitations and errors. As Table~\ref{tab:abstention_breakdown} shows, \texttt{missing} dominates abstentions across all methods; consequently, SpatialBench-UC partially measures \emph{detectability under COCO-trained detectors} rather than only spatial prompt following. Detector-level score filtering can also bound how checker thresholds take effect, if a detector discards boxes below its internal threshold, lowering the checker threshold below that will not recover those boxes. The human audit uses a single annotator and does not measure inter-annotator agreement. The benchmark scope is limited to two-object prompts and four axis-aligned relations. Finally, calibration uses the collected audit set and should be strengthened by larger, multi-annotator, and held-out calibration/validation protocols (future work).

\subsection{Conclusion}
SpatialBench-UC is a benchmark and evaluation toolkit for spatial prompt following under uncertainty. Instead of treating spatial compliance as a binary outcome, it explicitly represents uncertainty through abstention, calibrated confidence estimates, and risk-coverage reporting grounded in a lightweight human audit.

Beyond the fixed study reported here, the core contribution is a reusable harness designed for extension and reproducibility. We release a versioned benchmark bundle with structured metadata and provenance records that track prompts, generations, evaluator decisions, and calibration settings. Using this framework, we find that grounded generation strategies (BoxDiff and GLIGEN) substantially improve both pass rate and coverage over prompt-only baselines on our fixed runs, while also highlighting situations where the evaluator should abstain when the evidence is insufficient. We expect SpatialBench-UC to enable systematic comparisons across future models and settings by allowing users to swap generators, checkers, relations, and prompt templates while preserving consistent reporting and reproducibility guarantees.

\bibliographystyle{plainnat}
\bibliography{refs}

@misc{gokhale2022benchmarking,
  title        = {Benchmarking Spatial Relationships in Text-to-Image Generation},
  author       = {Gokhale, Tejas and Palangi, Hamid and Nushi, Besmira and Vineet, Vibhav and Horvitz, Eric and Kamar, Ece and Baral, Chitta and Yang, Yezhou},
  year         = {2022},
  eprint       = {2212.10015},
  archivePrefix= {arXiv},
  primaryClass = {cs.CV}
}

@inproceedings{ghosh2023geneval,
  title        = {GenEval: An Object-Focused Framework for Evaluating Text-to-Image Alignment},
  author       = {Ghosh, Dhruba and Hajishirzi, Hanna and Schmidt, Ludwig},
  booktitle    = {Advances in Neural Information Processing Systems (NeurIPS)},
  year         = {2023},
  eprint       = {2310.11513},
  archivePrefix= {arXiv},
  primaryClass = {cs.CV}
}

@inproceedings{huang2023t2icompbench,
  title        = {T2I-CompBench: A Comprehensive Benchmark for Open-world Compositional Text-to-image Generation},
  author       = {Huang, Kaiyi and Sun, Kaiyue and Xie, Enze and Li, Zhenguo and Liu, Xihui},
  booktitle    = {Advances in Neural Information Processing Systems (NeurIPS)},
  year         = {2023},
  volume       = {36},
  pages        = {78723--78747},
  eprint       = {2307.06350},
  archivePrefix= {arXiv},
  primaryClass = {cs.CV}
}

@InProceedings{Hu_2023_ICCV,
  author    = {Hu, Yushi and Liu, Benlin and Kasai, Jungo and Wang, Yizhong and Ostendorf, Mari and Krishna, Ranjay and Smith, Noah A.},
  title     = {TIFA: Accurate and Interpretable Text-to-Image Faithfulness Evaluation with Question Answering},
  booktitle = {Proceedings of the IEEE/CVF International Conference on Computer Vision (ICCV)},
  year      = {2023},
  pages     = {20406--20417}
}

@InProceedings{Thrush_2022_CVPR,
  author    = {Thrush, Tristan and Jiang, Ryan and Bartolo, Max and Singh, Amanpreet and Williams, Adina and Kiela, Douwe and Ross, Candace},
  title     = {Winoground: Probing Vision and Language Models for Visio-Linguistic Compositionality},
  booktitle = {Proceedings of the IEEE/CVF Conference on Computer Vision and Pattern Recognition (CVPR)},
  year      = {2022},
  pages     = {5238--5248},
  doi       = {10.1109/CVPR52688.2022.00517}
}

@InProceedings{Rombach_2022_CVPR,
  author    = {Rombach, Robin and Blattmann, Andreas and Lorenz, Dominik and Esser, Patrick and Ommer, Bj{\"o}rn},
  title     = {High-Resolution Image Synthesis With Latent Diffusion Models},
  booktitle = {Proceedings of the IEEE/CVF Conference on Computer Vision and Pattern Recognition (CVPR)},
  year      = {2022},
  pages     = {10674--10685},
  doi       = {10.1109/CVPR52688.2022.01042},
  eprint    = {2112.10752},
  archivePrefix = {arXiv},
  primaryClass  = {cs.CV}
}

@inproceedings{xie2023boxdiff,
  title        = {BoxDiff: Text-to-Image Synthesis with Training-Free Box-Constrained Diffusion},
  author       = {Xie, Jinheng and Li, Yuexiang and Huang, Yawen and Liu, Haozhe and Zhang, Wentian and Zheng, Yefeng and Shou, Mike Zheng},
  booktitle    = {Proceedings of the IEEE/CVF International Conference on Computer Vision (ICCV)},
  year         = {2023},
  pages        = {7452--7461},
  doi          = {10.1109/ICCV51070.2023.00685},
  eprint       = {2307.10816},
  archivePrefix= {arXiv},
  primaryClass = {cs.CV}
}

@inproceedings{li2023gligen,
  title        = {GLIGEN: Open-Set Grounded Text-to-Image Generation},
  author       = {Li, Yuheng and Liu, Haotian and Wu, Qingyang and Mu, Fangzhou and Yang, Jianwei and Gao, Jianfeng and Li, Chunyuan and Lee, Yong Jae},
  booktitle    = {Proceedings of the IEEE/CVF Conference on Computer Vision and Pattern Recognition (CVPR)},
  year         = {2023},
  pages        = {22511--22521},
  doi          = {10.1109/CVPR52729.2023.02156},
  eprint       = {2301.07093},
  archivePrefix= {arXiv},
  primaryClass = {cs.CV}
}

@InProceedings{Zhang_2023_ICCV_ControlNet,
  author    = {Zhang, Lvmin and Rao, Anyi and Agrawala, Maneesh},
  title     = {Adding Conditional Control to Text-to-Image Diffusion Models},
  booktitle = {Proceedings of the IEEE/CVF International Conference on Computer Vision (ICCV)},
  year      = {2023},
  pages     = {3836--3847},
  doi       = {10.1109/ICCV51070.2023.00355},
  note      = {arXiv:2302.05543}
}

@inproceedings{LinMBHPRDZ14,
  title     = {Microsoft {COCO}: Common Objects in Context},
  author    = {Lin, Tsung-Yi and Maire, Michael and Belongie, Serge and Hays, James and Perona, Pietro and Ramanan, Deva and Doll{\'a}r, Piotr and Zitnick, C. Lawrence},
  booktitle = {Computer Vision -- ECCV 2014},
  year      = {2014},
  pages     = {740--755},
  doi       = {10.1007/978-3-319-10602-1_48}
}

@inproceedings{ren2015faster,
  title     = {Faster {R-CNN}: Towards Real-Time Object Detection with Region Proposal Networks},
  author    = {Ren, Shaoqing and He, Kaiming and Girshick, Ross and Sun, Jian},
  booktitle = {Advances in Neural Information Processing Systems},
  year      = {2015},
  volume    = {28}
}

@article{liu2023grounding,
  title        = {Grounding {DINO}: Marrying {DINO} with Grounded Pre-Training for Open-Set Object Detection},
  author       = {Liu, Shilong and Zeng, Zhaoyang and Ren, Tianhe and Li, Feng and Zhang, Hao and Yang, Jie and Li, Chunyuan and Yang, Jianwei and Su, Hang and Zhu, Jun and Zhang, Lei},
  journal      = {arXiv preprint arXiv:2303.05499},
  year         = {2023}
}

@article{chow1970optimum,
  title   = {On optimum recognition error and reject tradeoff},
  author  = {Chow, C. K.},
  journal = {IEEE Transactions on Information Theory},
  volume  = {16},
  number  = {1},
  pages   = {41--46},
  year    = {1970},
  doi     = {10.1109/TIT.1970.1054406}
}

@inproceedings{GeifmanElYaniv2017Selective,
  author    = {Geifman, Yonatan and El-Yaniv, Ran},
  title     = {Selective Classification for Deep Neural Networks},
  booktitle = {Advances in Neural Information Processing Systems 30 (NeurIPS 2017)},
  year      = {2017},
  pages     = {4878--4887},
  eprint    = {1705.08500},
  archivePrefix = {arXiv},
  primaryClass  = {cs.LG}
}

@inproceedings{guo2017calibration,
  title     = {On Calibration of Modern Neural Networks},
  author    = {Guo, Chuan and Pleiss, Geoff and Sun, Yu and Weinberger, Kilian Q.},
  booktitle = {Proceedings of the 34th International Conference on Machine Learning (ICML)},
  year      = {2017},
  pages     = {1321--1330},
  eprint    = {1706.04599},
  archivePrefix= {arXiv},
  primaryClass = {cs.LG}
}

\clearpage
\appendix
\section{Reproducibility and additional results}
\subsection{Reproducibility package (exact artifact paths)}\label{app:repro_paths}
This paper is designed to be reproducible from a released benchmark bundle (fixed evaluation set). The canonical resources are:

\paragraph{Prompt dataset (versioned + hashed).}
\begin{itemize}
  \item \path{data/prompts/v1.0.1/prompts.jsonl}
  \item \path{data/prompts/v1.0.1/dataset_meta.json}
  \item \path{data/prompts/v1.0.1/sha256.txt}
\end{itemize}

\paragraph{Frozen generations and evaluator outputs.}
Run root:
\begin{itemize}
  \item \path{runs/final_20260112_084335/}
\end{itemize}
Subruns:
\begin{itemize}
  \item \path{runs/final_20260112_084335/sd15_promptonly/}
  \item \path{runs/final_20260112_084335/sd15_boxdiff/}
  \item \path{runs/final_20260112_084335/sd14_gligen/}
\end{itemize}
Per subrun (calibrated checker outputs):
\begin{itemize}
  \item \path{manifest.jsonl}
  \item \path{eval/per_sample.jsonl}
  \item \path{eval/metrics.json}
  \item \path{eval/provenance.json}
  \item \path{eval/checker_config.yaml}
  \item \path{eval/overlays/*.png} (optional; large; not required to reproduce tables)
\end{itemize}
Uncalibrated outputs used to reproduce the uncalibrated report are stored as:
\begin{itemize}
  \item \path{eval_precal_20260116_113552/per_sample.jsonl}
  \item \path{eval_precal_20260116_113552/metrics.json}
  \item \path{eval_precal_20260116_113552/checker_config.yaml}
\end{itemize}

\paragraph{Reports (uncalibrated vs calibrated).}
\begin{itemize}
  \item Uncalibrated report: \path{runs/final_20260112_084335/reports/v1_finalfix_20260114_143137/}
  \item Calibrated report: \path{runs/final_20260112_084335/reports/v1_calibrated_20260116_113552/}
\end{itemize}
Each report includes \path{tables/*.csv}, \path{assets/*.png}, and provenance in \path{report_meta.json}. For reproducibility, \path{report_config_effective.yaml} records the resolved run list and evaluation subdirectory (the copied \path{report_config.yaml} may be a template and should not be treated as authoritative for the run list).

\paragraph{Human audit and calibration artifacts.}
\begin{itemize}
  \item Sample definition: \path{audits/v1/sample.csv}
  \item Human labels: \path{audits/v1/labels_filled.json} (and \path{.csv})
  \item Baseline analysis: \path{audits/v1/analysis_uncalibrated/}
  \item Calibrated analysis: \path{audits/v1/analysis_calibrated/}
  \item Calibration selection metadata (grid search outputs): \path{audits/v1/analysis_calibration/}
\end{itemize}
Reproducing the calibration grid search itself requires access to image artifacts under \path{runs/**/images/}; the released \path{audit_metrics.json} records the selected parameters without requiring regeneration.

\paragraph{Configs and entrypoints.}
\begin{itemize}
  \item Final checker config: \path{configs/checker_v1.yaml} (backup: \path{configs/checker_v1.backup_20260116_113201.yaml})
  \item Generator config: \path{configs/gen_sd15_promptonly.yaml}
  \item Generator config: \path{configs/gen_sd15_boxdiff.yaml}
  \item Generator config: \path{configs/gen_sd15_gligen.yaml}
  \item Entrypoint: \path{src/spatialbench_uc/generate.py}
  \item Entrypoint: \path{src/spatialbench_uc/evaluate.py}
  \item Entrypoint: \path{src/spatialbench_uc/report.py}
  \item Entrypoint: \path{src/spatialbench_uc/audit/sample.py}
  \item Entrypoint: \path{src/spatialbench_uc/audit/analyze.py}
\end{itemize}

\subsection{Additional quantitative results and plots}\label{app:additional_results}
This appendix collects additional tables and plots referenced in the main text, including uncalibrated results and calibration deltas.
\begin{table}[H]
  \centering
  \begin{tabular}{@{}lrrrrr@{}}
\toprule
Method & Pass (\%) & Coverage (\%) & Pass$\mid$Decided (\%) & PromptPass (\%) & MeanConf \\
\midrule
SD1.5 Prompt-only & 14.5 & 28.4 & 51.1 & 40.5 & 0.246 \\
SD1.5 BoxDiff & 41.1 & 43.4 & 94.8 & 76.5 & 0.410 \\
SD1.4 GLIGEN & 51.8 & 52.1 & 99.3 & 78.5 & 0.507 \\
\bottomrule
\end{tabular}

  \caption{Main results (uncalibrated checker).}
  \label{tab:main_results_uncal}
\end{table}

\begin{table}[H]
  \centering
  \resizebox{\linewidth}{!}{\begin{tabular}{@{}lrrrrr@{}}
\toprule
Method & $\Delta$Pass (pp) & $\Delta$Coverage (pp) & $\Delta$Pass$\mid$Decided (pp) & $\Delta$PromptPass (pp) & $\Delta$MeanConf \\
\midrule
SD1.5 Prompt-only & -2.75 & -4.63 & -1.63 & -6.50 & -0.040 \\
SD1.5 BoxDiff & -0.76 & -0.88 & +0.19 & -0.50 & -0.015 \\
SD1.4 GLIGEN & -0.13 & -0.12 & +0.00 & +0.00 & -0.001 \\
\bottomrule
\end{tabular}

}
  \caption{Calibration deltas (calibrated minus uncalibrated).}
  \label{tab:main_results_delta}
\end{table}

\begin{table}[H]
  \centering
  \begin{tabular}{@{}lrrrr@{}}
\toprule
Method & Above (\%) & Below (\%) & Left-of (\%) & Right-of (\%) \\
\midrule
SD1.5 Prompt-only & 16.5 & 14.0 & 7.5 & 9.0 \\
SD1.5 BoxDiff & 42.0 & 38.5 & 40.0 & 41.0 \\
SD1.4 GLIGEN & 52.5 & 49.0 & 51.5 & 53.5 \\
\bottomrule
\end{tabular}

  \caption{Pass rate by relation (calibrated report).}
  \label{tab:by_relation_calibrated}
\end{table}

\begin{table}[H]
  \centering
  \begin{tabular}{@{}lrrrr@{}}
\toprule
Method & Pairs & Both-pass (\%) & Undecidable (\%) & One-sided (\%) \\
\midrule
SD1.5 Prompt-only & 100 & 19.0 & 81.0 & 0.0 \\
SD1.5 BoxDiff & 100 & 66.0 & 34.0 & 0.0 \\
SD1.4 GLIGEN & 100 & 74.0 & 26.0 & 0.0 \\
\bottomrule
\end{tabular}

  \caption{Counterfactual outcomes (calibrated report).}
  \label{tab:counterfactual_calibrated}
\end{table}

\begin{figure}[H]
  \centering
  \includegraphics[width=0.65\linewidth]{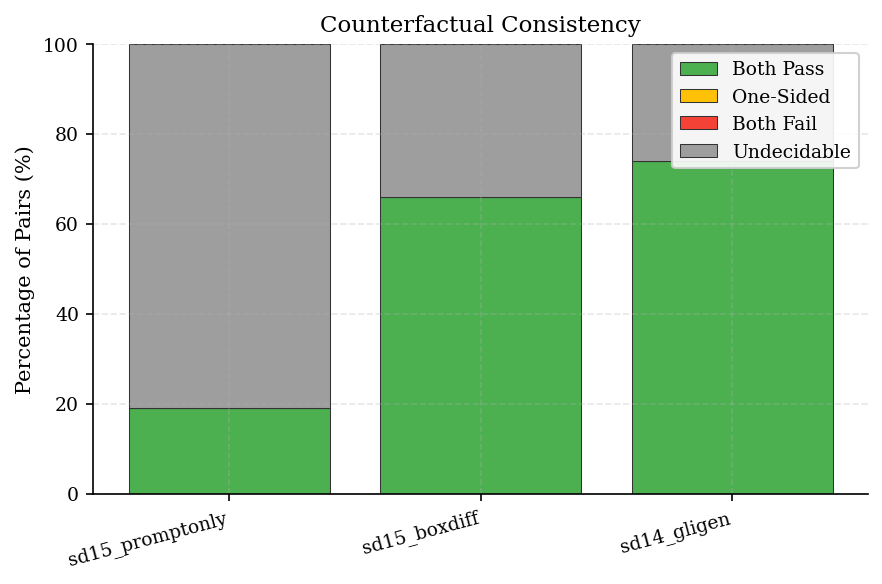}
  \caption{Counterfactual consistency (calibrated report), both-pass rate and undecidable mass over paired prompts.}
  \label{fig:counterfactual_consistency}
\end{figure}

\begin{figure}[H]
  \centering
  \includegraphics[width=0.65\linewidth]{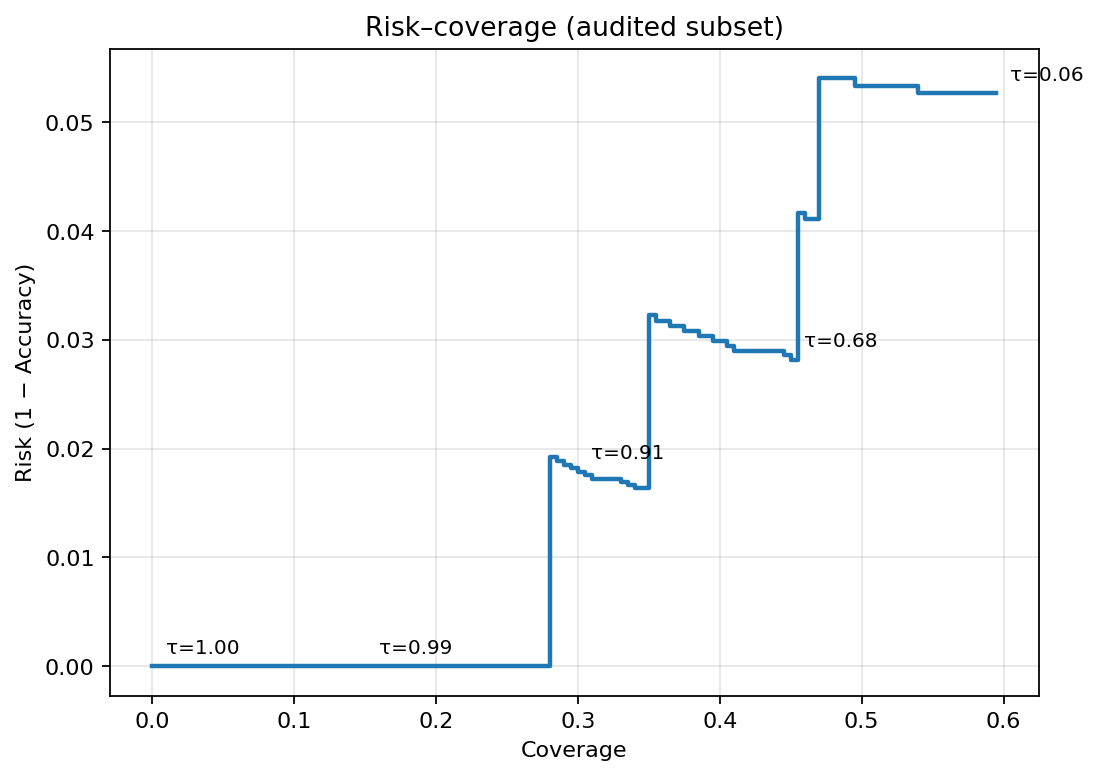}
  \caption{Risk--coverage curve for the uncalibrated checker (audited subset; sweep over unique confidence values). The curve is step-like because coverage changes only when $\tau$ crosses a sample’s discrete confidence value.}
  \label{fig:risk_coverage_uncal}
\end{figure}

\begin{figure}[H]
  \centering
  \includegraphics[width=0.65\linewidth]{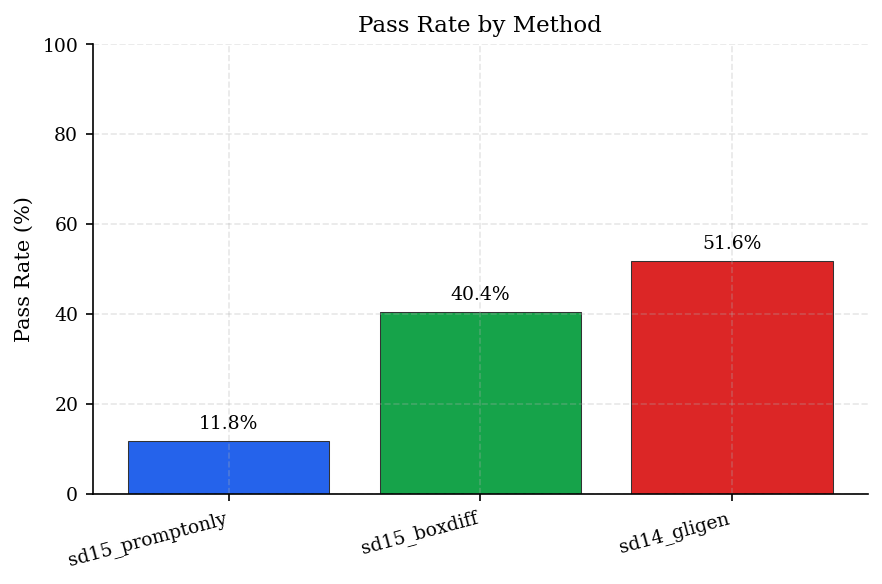}
  \caption{Pass rate comparison (calibrated report).}
  \label{fig:pass_rate_comparison}
\end{figure}

\begin{figure}[H]
  \centering
  \includegraphics[width=0.65\linewidth]{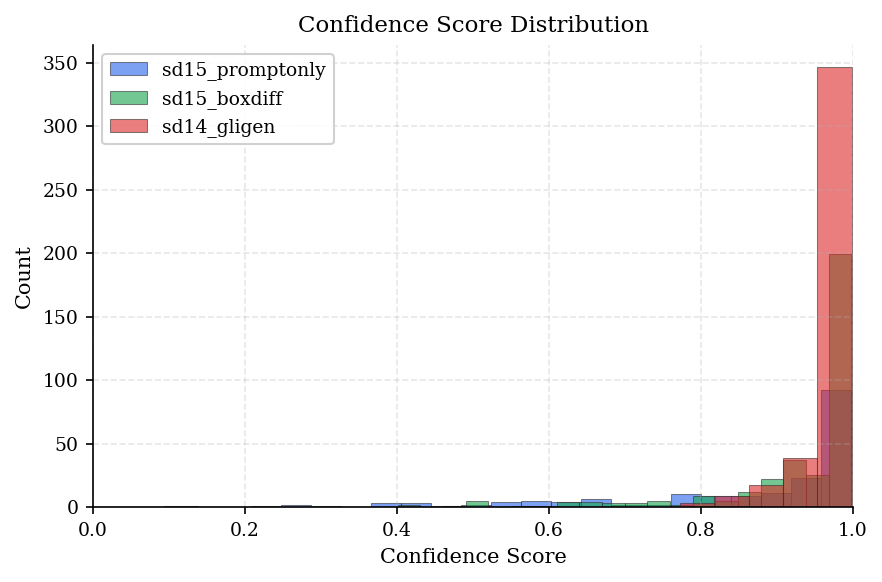}
  \caption{Confidence distribution (calibrated report).}
  \label{fig:confidence_distribution}
\end{figure}

\subsection{Qualitative gallery (additional examples)}\label{app:qual_gallery}
\begin{figure}[H]
  \centering
  \begin{subfigure}[t]{0.49\linewidth}
    \centering
    \includegraphics[width=\linewidth]{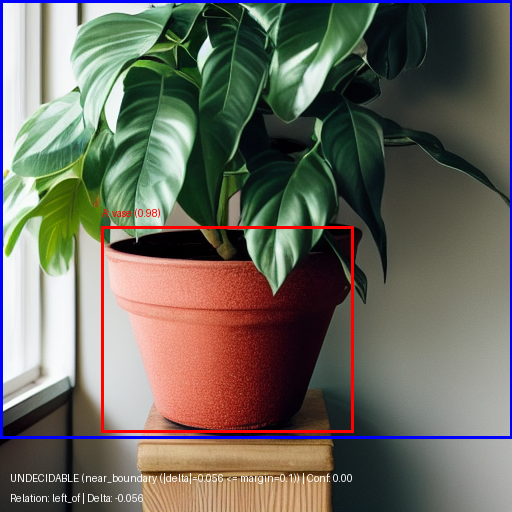}
    \caption{UNDECIDABLE (checker: near-boundary; human: no vase)\\``vase left of potted plant''}
  \end{subfigure}\hfill
  \begin{subfigure}[t]{0.49\linewidth}
    \centering
    \includegraphics[width=\linewidth]{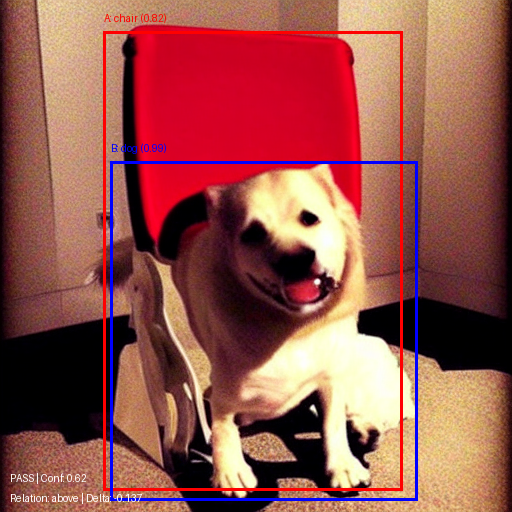}
    \caption{False PASS (BoxDiff)\\``chair above dog''}
  \end{subfigure}

  \vspace{4pt}

  \begin{subfigure}[t]{0.49\linewidth}
    \centering
    \includegraphics[width=\linewidth]{figures/overlay_promptonly_overlap.png}
    \caption{UNDECIDABLE (near-boundary)\\``chair left of dog''}
  \end{subfigure}\hfill
  \begin{subfigure}[t]{0.49\linewidth}
    \centering
    \includegraphics[width=\linewidth]{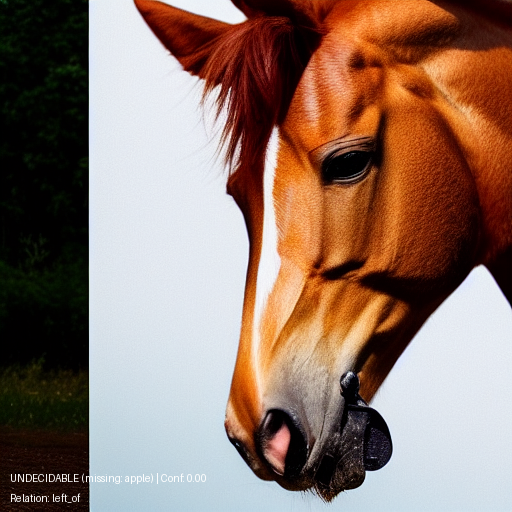}
    \caption{UNDECIDABLE (missing detection)\\``apple left of horse''}
  \end{subfigure}
  \caption{Qualitative gallery: representative incorrect decisions and abstentions. The overlay text shows the checker verdict and abstention reason.}
  \label{fig:qual_gallery}
\end{figure}

\end{document}